\documentclass[11pt]{article}

\usepackage[final]{acl}

\usepackage{times}
\usepackage{latexsym}
\usepackage{float}
\usepackage{graphicx}
\usepackage[T1]{fontenc}
\usepackage[utf8]{inputenc}
\usepackage{microtype}
\usepackage{inconsolata}
\usepackage{tabularx}
\usepackage{booktabs}
\usepackage{amsmath}
\usepackage{orcidlink}

\title{ShriNep@EEUCA 2026: RAKSHAK -- Multi-Task DeBERTa with Rationale Distillation and Jigsaw-Augmented Training for Toxic Intent Classification}

\author{
  Binayak Karki$^1$\orcidlink{0009-0001-2386-0578} \quad Aryan Kafle$^2$\orcidlink{0009-0007-2251-2671} \quad Pingala Ghimire$^3$\orcidlink{0009-0008-6183-9834} \\[4pt]
  $^1$ Mechi Multiple Campus, Nepal \\
  $^2$ Northern Kentucky University, USA \\
  $^3$ Himalaya College of Engineering, Nepal \\[4pt]
  \texttt{binayak.805421@memc.tu.edu.np}, \texttt{kaflea3@mymail.nku.edu}, \texttt{pingalaghimire555@gmail.com}
}

\begin{document}
\maketitle

\begin{abstract}
This paper presents two systems for the GameTox Shared Task at the Workshop on EEUCA at ACL 2026, which requires classifying World of Tanks chat utterances into six fine-grained toxic intent categories (Labels 0--5). Severe class imbalance, domain-specific multilingual slang, and extremely scarce data for rare categories such as Threats (Label~4, 60 samples) and Extremism (Label~5, 24 samples) make this a challenging classification problem. Our primary submission, RAKSHAK (rak\d{s}aka, Sanskrit for ``Protector''), is a multi-task DeBERTa-v3-base \citep{he2021debertav3} framework combining rationale distillation from Qwen2.5-14B \citep{qwen25}, Supervised Contrastive Loss, and dedicated rare-class binary heads. RAKSHAK's training data is augmented with cross-domain transfer from the Jigsaw Toxic Comment dataset (16,225 samples mapped to Labels 1--4) and 100 LLM-generated extremism samples for Label~5. Our secondary system (M1) fine-tunes DeBERTa-v3-base with Focal Loss on the original GameTox data plus the same 100 extremism samples, without Jigsaw transfer. RAKSHAK achieves a Macro F1 of \textbf{0.5883} on the official test set, ranking \textbf{7th out of 35} participating teams, while M1 achieves 0.5252 Macro F1. An ablation comparing M1 with and without Jigsaw data shows that cross-domain transfer accounts for +2.6 F1 points, while RAKSHAK's multi-task architecture contributes a further +3.7 points.
\end{abstract}

\section{Introduction}

Online multiplayer games rely on in-game chat for coordination, yet these channels also carry harmful content ranging from profanity to extremist material \citep{parihar2021hate}. Automatic moderation matters for player safety, but game chat is noisy, multilingual, and heavily skewed toward non-toxic messages, making reliable classification difficult \citep{thapa2025large}.

The GameTox Shared Task at EEUCA 2026 \citep{hurriyetouglu2026eeuca, thapa2026toxicity} evaluates this challenge on approximately 53,000 World of Tanks utterances annotated into six intent labels (0--5), from non-toxic to extremism. Systems are ranked by Macro F1, placing strong emphasis on performance across all classes, including those with very few training samples.

Prior work on toxicity detection has largely focused on social media \citep{waseem2016hateful, davidson2017automated} and transfers poorly to gaming language, where jargon, obfuscation, and code-switching are common. Large-scale annotation efforts like the Jigsaw dataset \citep{jigsaw2018toxic} showed the value of cross-domain data, but the social media register differs sharply from gaming chat. On the modelling side, knowledge distillation from large LLMs \citep{hinton2015distilling, hsieh2023distilling, magister2023teaching}, Focal Loss for class imbalance \citep{lin2017focal}, and supervised contrastive learning \citep{khosla2020supervised} have all shown promise; chain-of-thought rationales \citep{wei2022chain} further suggest that structured teacher explanations transfer reasoning that labels alone cannot.

We draw on these techniques in two systems: \textbf{RAKSHAK} (primary), a multi-task DeBERTa-v3-base framework combining rationale distillation from Qwen2.5-14B, Supervised Contrastive Loss, rare-class binary heads, and two-stage augmentation via Jigsaw transfer and LLM-generated extremism samples; and \textbf{M1} (secondary), a DeBERTa-v3-base model fine-tuned with Focal Loss on a smaller augmented set. Beyond gaming, LLMs are now used in settings where misclassification carries real consequences, from clinical diagnosis \citep{yan2025llm} to museum visitor assistance \citep{guragain2025personalized}, making reliable content moderation a concern well beyond this domain.

\section{Related Work}

\paragraph{Toxicity detection.} Early approaches to online toxicity detection relied on feature-engineered classifiers \citep{waseem2016hateful, davidson2017automated}, while recent work has shifted toward fine-tuning pretrained language models on curated datasets. Ensemble methods combining multiple multilingual BERT-based models have shown strong results on shared task benchmarks for hate speech detection, with data augmentation and class-imbalance handling being key contributors to performance \citep{guragain2025nlpineers}. Most existing research targets social media platforms, and relatively few studies address the distinct challenges of gaming environments, where language is heavily obfuscated, multilingual, and laden with domain-specific slang \citep{parihar2021hate}. The GameTox dataset \citep{naseem2025gametox} is among the first large-scale resources specifically targeting gaming chat toxicity.

\paragraph{Knowledge distillation and rationale augmentation.} \citet{hinton2015distilling} introduced knowledge distillation via soft logit matching between teacher and student models. More recently, \citet{hsieh2023distilling} proposed distilling step-by-step, where a large teacher generates natural language rationales that are concatenated with inputs during student training, enabling small models to outperform larger ones with less data. \citet{magister2023teaching} and \citet{li2023symbolic} demonstrated similar rationale distillation approaches for teaching reasoning to small language models. Our RAKSHAK framework follows this paradigm, using Qwen2.5-14B \citep{qwen25} as the teacher to generate structured rationales for a DeBERTa-v3-base \citep{he2021debertav3} student.

\paragraph{Contrastive learning for text classification.} Supervised Contrastive Loss \citep{khosla2020supervised} has been shown to improve representation quality by pulling same-class embeddings together while pushing apart different-class embeddings. This is particularly beneficial under class imbalance, as rare-class samples receive stronger gradient signal through explicit pairwise comparisons rather than relying solely on cross-entropy with the majority class.

\paragraph{Loss reweighting for imbalanced classification.} Focal Loss \citep{lin2017focal}, originally proposed for object detection, down-weights well-classified examples to focus training on hard cases. It has since been widely adopted for imbalanced text classification tasks, including toxicity detection, where the dominant non-toxic class can overwhelm standard cross-entropy training.

\section{Background and Task Description}

\subsection{Task Setup}
This Shared Task is organised within the 9th Workshop on Event Extraction and Understanding: Challenges and Applications (EEUCA) at ACL 2026 \citep{hurriyetouglu2026eeuca}, under the CHiPSAL track. The task focuses on intent classification of in-game chat utterances from World of Tanks, with a globally distributed multilingual player base.

Given an utterance $u$ from the game chat, systems must predict an intent label $y \in \{0, 1, 2, 3, 4, 5\}$ as defined in Table~\ref{tab:label-taxonomy}.

\begin{table}[h!]
  \centering
  \small
  \begin{tabularx}{\columnwidth}{@{} c l X @{}}
    \toprule
    \textbf{Label} & \textbf{Category} & \textbf{Description \& Example} \\
    \midrule
    0 & Non-toxic
      & Benign communication, strategy, or neutral chatter. \smallskip\newline \textit{``good game''} \\
    1 & Insults \& Flaming
      & Personal attacks or profanity directed at other players. \smallskip\newline \textit{``fuckin noob''} \\
    2 & Other Offensive
      & Offensive content not fitting other categories. \smallskip\newline \textit{``learm to p\textasciicircum ay stuopid red player''} \\
    3 & Hate \& Harassment
      & Identity-based hate or sustained harassment. \smallskip\newline \textit{``a fckng dum bstrd from easteurope''} \\
    4 & Threats
      & Direct or implicit threats of harm or violence. \smallskip\newline \textit{``hope your family die in fire''} \\
    5 & Extremism
      & Extremist content, radicalisation, or incitement. \smallskip\newline \textit{``STUYOU RUSIA AND NAZ1 LEMMIN O ALL ONE SIDE''} \\
    \bottomrule
  \end{tabularx}
  \caption{Toxicity label taxonomy for the EEUCA 2026 shared task \citep{thapa2026toxicity, naseem2025gametox}.}
  \label{tab:label-taxonomy}
\end{table}

\subsection{Dataset}

The GameTox dataset \citep{naseem2025gametox} comprises approximately 53,000 utterances across train, validation, and test splits from World of Tanks in-game chat logs, with 42,959 samples in the training set. The annotation schema is adapted from the CrisisHateMM framework \citep{bhandari2023crisishatemm}. The dataset exhibits extreme class imbalance: Label~0 (Non-toxic) accounts for over 80\% of training data, while Label~5 (Extremism) has only 24 samples and Label~4 (Threats) has 60. The corpus is multilingual, containing utterances predominantly in English alongside Polish, Russian, German, French, and other languages reflecting the worldwide player base.

\section{System Description}
\label{sec:system_description}

\subsection{Data Augmentation}
\label{sec:augmentation}

We employ two data augmentation strategies targeting underrepresented toxic classes, following the broader observation that augmentation is critical for rare-class performance in hate speech shared tasks \citep{guragain2025nlpineers}. Both strategies are used for RAKSHAK; M1 uses only the LLM-generated extremism samples (Section~\ref{sec:extremism-aug}).

\subsubsection{Cross-Domain Transfer from Jigsaw}
\label{sec:jigsaw}

To enrich the scarce in-domain toxic samples for RAKSHAK, we incorporate data from the Jigsaw Toxic Comment Classification dataset \citep{jigsaw2018toxic}, mapping its multi-label toxicity annotations to the GameTox intent taxonomy as shown in Table~\ref{tab:jigsaw-mapping}. To validate the mapping, we sampled 10 examples from each Jigsaw category and independently prompted two LLMs (Gemini 1.5 Pro and Grok) to assign GameTox labels; both models agreed on the same mapping for all categories. The Jigsaw dataset also contains a large non-toxic category which maps naturally to GameTox Label~0; however, we exclude these samples since Label~0 is already heavily overrepresented. No suitable Jigsaw category exists for Label~5 (Extremism). For samples with multiple active Jigsaw labels, we assign the highest-severity GameTox label (e.g., a sample tagged both \texttt{obscene} and \texttt{threat} is mapped to Label~4). After mapping and deduplication, this yields 16,225 additional samples across Labels 1--4.

\begin{table}[h!]
  \centering
  \small
  \begin{tabularx}{\columnwidth}{@{} l c r @{}}
    \toprule
    \textbf{Jigsaw Label} & \textbf{GameTox} & \textbf{Samples} \\
    \midrule
    toxic, obscene, insult        & Label 1   & 6{,}500 \\
    other\_offensive              & Label 2   & 7{,}940 \\
    severe\_toxic, identity\_hate & Label 3   & 1{,}307 \\
    threat                        & Label 4   & 478 \\
    \midrule
    non-toxic                     & Label 0   & \textit{excluded} \\
    (no mapping)                  & Label 5   & --- \\
    \bottomrule
  \end{tabularx}
  \caption{Mapping from Jigsaw labels to GameTox categories (see Table~\ref{tab:label-taxonomy}). Non-toxic samples are excluded. No Jigsaw category maps to Extremism.}
  \label{tab:jigsaw-mapping}
\end{table}

\subsubsection{LLM-Generated Extremism Samples}
\label{sec:extremism-aug}

Label~5 (Extremism) has no Jigsaw counterpart, leaving only 24 in-domain training samples. We generate 100 synthetic extremism samples through a four-step pipeline:

\begin{enumerate}
    \item \textbf{Keyword mining:} Extract extremism-relevant keywords (slurs, political references, radicalisation terms) from the existing Label~5 training samples.
    \item \textbf{Keyword expansion:} Use Grok to produce morphological variants, obfuscated spellings, and semantically related terms, expanding the seed keyword list.
    \item \textbf{Sentence generation:} Prompt Qwen2.5-14B \citep{qwen25} with a task-specific instruction describing the GameTox shared task and the definition of extremism from \citet{naseem2025gametox}. To work around safety filters, extremist keywords are replaced with placeholder tokens (e.g., \texttt{[WORD1]}, \texttt{[WORD2]}) in the prompt, and Qwen generates sentence frames containing these placeholders. Qwen2.5-14B was selected as the strongest open-weight model that could be served locally via Ollama within our compute constraints, supporting reproducibility without dependence on closed-source APIs. The prompt template is provided in Appendix~\ref{sec:prompt}.
    \item \textbf{Keyword injection:} Replace placeholder tokens in generated sentences with real extremist keywords from Steps~1 and 2.
\end{enumerate}

This pipeline addresses the dual challenge of data scarcity and LLM safety refusal when generating harmful content for research purposes. The 100 extremism samples are used by both M1 and RAKSHAK. We note that these samples were not formally verified for exact-string overlap with the official test set; this is acknowledged in our limitations.

Table~\ref{tab:data-composition} summarises the training data composition for each system.

\begin{table}[t]
  \centering
  \small
  \setlength{\tabcolsep}{3pt}
  \begin{tabularx}{\columnwidth}{@{} l c r r r @{}}
    \toprule
    \textbf{Category} & \textbf{L} & \textbf{Original} & \textbf{M1} & \textbf{RAKSHAK} \\
    \midrule
    Non-toxic       & 0 & 34{,}797 & 34{,}797 & 34{,}797 \\
    Insults         & 1 & 5{,}925  & 5{,}925  & 12{,}425 \\
    Other Offensive & 2 & 1{,}874  & 1{,}874  & 9{,}814  \\
    Hate \& Harass. & 3 & 279      & 279      & 1{,}586  \\
    Threats         & 4 & 60       & 60       & 538      \\
    Extremism       & 5 & 24       & 124      & 124      \\
    \midrule
    \textbf{Total}  &   & 42{,}959 & 43{,}059 & 59{,}284 \\
    \bottomrule
  \end{tabularx}
  \caption{Training data composition. M1 uses the original GameTox data plus 100 LLM-generated extremism samples. RAKSHAK additionally incorporates 16,225 Jigsaw-transferred samples across Labels 1--4.}
  \label{tab:data-composition}
\end{table}

\subsection{M1: DeBERTa-v3-base with Focal Loss}
\label{sec:m1}

Our secondary system fine-tunes DeBERTa-v3-base \citep{he2021debertav3} as a single-stage six-class intent classifier on the original GameTox training data plus 100 LLM-generated extremism samples (Section~\ref{sec:extremism-aug}). The classification head is a linear layer over the \texttt{[CLS]} representation producing 6-class logits, trained with Focal Loss \citep{lin2017focal} ($\gamma = 2.0$) to down-weight well-classified majority-class examples and direct gradient updates toward hard, minority-class samples. The model is trained for 5 epochs with a learning rate of 2e-5, batch size of 32, and gradient clipping at 1.0. Model selection is based on the best validation Macro F1 on a 90/10 train-validation split (seed=42). At inference, the model predicts directly among all six labels in a single forward pass.

\subsection{RAKSHAK: Multi-Task Rationale Distillation Framework}
\label{sec:rakshak}

\begin{figure*}[t]
    \centering
    \includegraphics[width=0.95\textwidth]{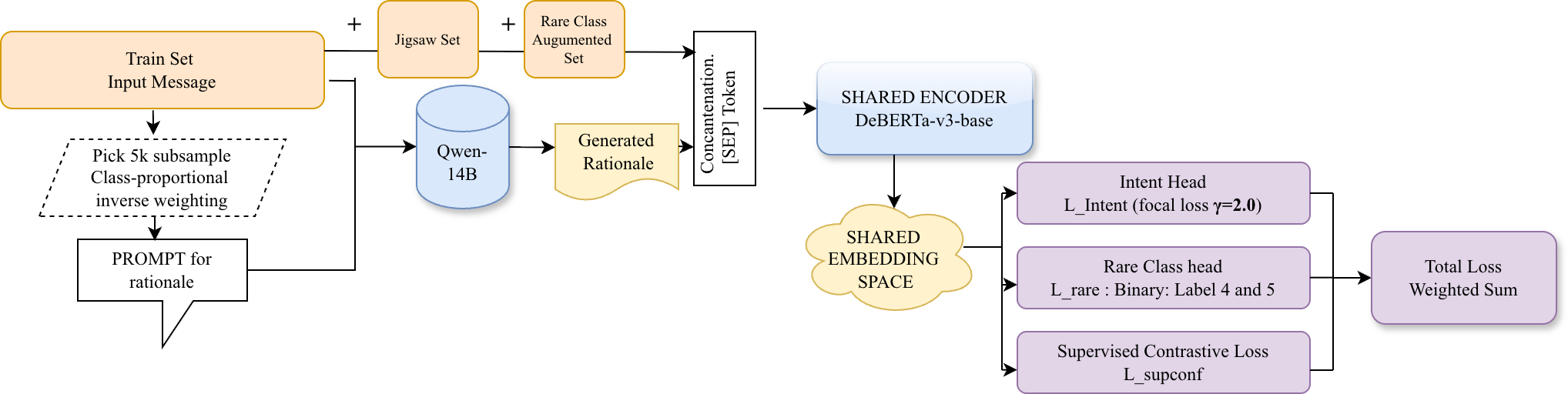}
    \caption{RAKSHAK architecture. The shared DeBERTa-v3-base encoder receives concatenated input messages and teacher-generated rationales. Three loss components operate over the shared embedding space: Focal Loss on the 6-class intent head, binary cross-entropy on dedicated Label~4 and Label~5 heads (weighted $2\times$), and Supervised Contrastive Loss on the \texttt{[CLS]} embeddings.}
    \label{fig:rakshak_arch}
\end{figure*}

RAKSHAK is our primary system. It extends the DeBERTa-v3-base backbone into a multi-task learning framework that addresses class imbalance through three mechanisms: (1)~rationale-augmented knowledge distillation from a teacher LLM, following the distill-then-train paradigm of \citet{hsieh2023distilling}, (2)~dedicated rare-class binary classifiers, and (3)~Supervised Contrastive Loss \citep{khosla2020supervised} on the shared embedding space. Unlike M1, RAKSHAK trains on the full augmented dataset including Jigsaw-transferred samples (Table~\ref{tab:data-composition}). Training proceeds in two phases: Phase~1 generates natural language rationales using Qwen2.5-14B \citep{qwen25}, and Phase~2 trains the student encoder on rationale-augmented inputs under the combined multi-task loss. Figure~\ref{fig:rakshak_arch} presents the architecture.

\subsubsection{Teacher Rationale Generation}

Qwen2.5-14B (served locally via Ollama, temperature 0.3, top-p 0.9, max 200 tokens) generates a structured explanation for each selected training sample. Each rationale identifies toxic keywords, provides gaming-specific context, infers intent, and assigns the corresponding GameTox category. An example:
\begin{quote}
    \textit{``This message contains `kurwa' (Polish profanity) and `uninstall' (a gaming-specific threat), indicating Label~1 (Insults and Flaming). Intent: demeaning a teammate. Category: Toxic.''}
\end{quote}

We select 5{,}000 training samples for rationale generation using class-proportional inverse weighting, allocating more rationales to rarer classes relative to their natural frequency. This directs the majority of the generation budget toward Labels 3--5 where the model most benefits from additional reasoning signal, while spending less compute on the well-represented majority class. Rationales are saved incrementally every 50 samples to support resumption after interruptions.

\subsubsection{Rationale-Augmented Training}

Rather than distilling soft logits from the teacher, RAKSHAK concatenates the teacher's rationale directly to the input before tokenisation:

\medskip
\noindent\texttt{[MESSAGE] [SEP] [RATIONALE: ...]}
\medskip

\noindent This is motivated by \citet{hsieh2023distilling}, who showed that natural language rationales can transfer reasoning from a large teacher to a small student more effectively than logit-based distillation. The concatenated input is tokenised and truncated to 128 tokens, accommodating both the original message and most of each rationale.

Rationales are used exclusively during training; at inference, the model receives only the raw message. This is a deliberate design choice: serving a 14B-parameter teacher at inference would negate the efficiency advantage of the student encoder, and we treat the rationale as privileged training context whose benefit is expected to persist in the learned representations at test time. We acknowledge that this introduces a train/test input distribution shift, as the encoder is optimised on inputs of the form \texttt{[MESSAGE] [SEP] [RATIONALE]} but evaluated on \texttt{[MESSAGE]} alone; the implications of this are discussed in the Limitations section.

\subsubsection{Multi-Task Heads}

Two types of classification heads are trained jointly over the shared \texttt{[CLS]} representation:

\begin{itemize}
    \item \textbf{Intent Head (primary):} A two-layer MLP (768 $\rightarrow$ 256 $\rightarrow$ ReLU $\rightarrow$ Dropout $\rightarrow$ 6 logits), trained with Focal Loss ($\gamma=2.0$). This head produces all final predictions at inference.
    \item \textbf{Rare-Class Heads:} Two independent binary classifiers (each a two-layer MLP), one for Label~4 (Threats) and one for Label~5 (Extremism). Their losses are summed and weighted $2.0\times$ in the total objective. These heads serve as auxiliary training signals that encourage the shared encoder to develop representations discriminative for the rarest categories.
\end{itemize}

At inference, only the intent head is used. The rare-class heads contribute exclusively during training by shaping the shared representation.

\subsubsection{Loss Function}

The total training objective combines three components:
\begin{equation}
\mathcal{L}_{\text{total}} = \mathcal{L}_{\text{focal}} + 0.3 \cdot \mathcal{L}_{\text{supcon}} + 2.0 \cdot (\mathcal{L}_{\text{L4}} + \mathcal{L}_{\text{L5}})
\end{equation}

\noindent \textbf{Focal Loss} \citep{lin2017focal} on the intent head:
\begin{equation}
FL(p_t) = -\alpha_t(1-p_t)^\gamma \log(p_t)
\end{equation}
with $\gamma=2.0$, focusing training on hard examples by down-weighting well-classified samples.

\noindent \textbf{Supervised Contrastive Loss} \citep{khosla2020supervised} operates directly on the \texttt{[CLS]} embeddings:
\begin{equation}
\mathcal{L}_{\text{supcon}} = -\log\!\left[
  \frac{
    \exp\!\bigl(\mathrm{sim}(z_i, z_p)/\tau\bigr)
  }{
    \displaystyle\sum_{j \neq i} \exp\!\bigl(\mathrm{sim}(z_i, z_j)/\tau\bigr)
  }
\right]
\end{equation}
With temperature $\tau=0.07$, this pulls same-class embeddings into tighter clusters in the shared representation space, especially beneficial for the rarest classes.

\noindent \textbf{Rare-class binary losses} ($\mathcal{L}_{\text{L4}}$, $\mathcal{L}_{\text{L5}}$) are standard binary cross-entropy on the dedicated heads, weighted $2.0\times$ to ensure that gradients from rare classes exert sufficient influence on the shared encoder. Table~\ref{tab:rakshak_hyperparams} summarises the complete training configuration.

\begin{table}[t]
\centering
\scriptsize
\setlength{\tabcolsep}{3pt}
\begin{tabularx}{\columnwidth}{@{}X >{\raggedleft\arraybackslash}p{0.36\columnwidth}@{}}
\toprule
\textbf{Hyperparameter} & \textbf{Value} \\
\midrule
Backbone & DeBERTa-v3-base \\
Max sequence length & 128 tokens \\
Batch size / LR / Epochs & 32 / 2e-5 / 5 \\
LR schedule & Linear warmup (10\%), clip 1.0 \\
Train / val split & 90\% / 10\% (seed=42) \\
Rationale teacher & Qwen2.5-14B (Ollama) \\
Rationale samples & 5{,}000 (inverse-weighted) \\
Intent loss (Focal, $\gamma=2.0$) & weight 1.0 \\
Rare-class loss (L4 + L5) & weight $2.0\times$ \\
SupCon weight / $\tau$ & 0.3 / 0.07 \\
Model selection & Best val Macro F1 \\
\bottomrule
\end{tabularx}
\caption{Hyperparameters and loss configuration for RAKSHAK.}
\label{tab:rakshak_hyperparams}
\end{table}

\section{Results and Discussion}

Table~\ref{tab:results} presents the official test-set results. RAKSHAK achieved a Macro F1 of \textbf{0.5883}, ranking \textbf{7th out of 35} teams on the shared task leaderboard \citep{thapa2026toxicity}. M1 achieved a Macro F1 of 0.5252.

\begin{table}[t]
\centering
\small
\setlength{\tabcolsep}{3pt}
\begin{tabularx}{\columnwidth}{@{} l
    >{\centering\arraybackslash}X
    >{\centering\arraybackslash}X
    >{\centering\arraybackslash}X @{}}
\toprule
\textbf{Metric} & \textbf{M1} & \textbf{M1+Jigsaw} & \textbf{RAKSHAK} \\
\midrule
F1 Macro           & 0.5252 & 0.5512 & \textbf{0.5883} \\
Accuracy           & 0.8147 & 0.8930 & \textbf{0.9031} \\
Precision (Macro)  & 0.4882 & 0.5201 & \textbf{0.5540} \\
Recall (Macro)     & 0.6482 & 0.6476 & \textbf{0.6590} \\
\bottomrule
\end{tabularx}
\caption{Official test-set results. M1 trains on GameTox + 100 extremism samples. M1+Jigsaw adds Jigsaw transfer. RAKSHAK adds the multi-task architecture on top of M1+Jigsaw data. RAKSHAK is the primary submission (ranked 7th/35).}
\label{tab:results}
\end{table}

RAKSHAK outperforms M1 across all reported metrics, with a Macro F1 advantage of over 6 points. To disentangle the contributions of data augmentation and architecture, we additionally evaluate M1 trained with the same Jigsaw-augmented data as RAKSHAK (M1+Jigsaw in Table~\ref{tab:results}). The breakdown is clear: Jigsaw transfer alone improves M1 from 0.5252 to 0.5512 (+2.6 points), while RAKSHAK's multi-task architecture adds a further 3.7 points on top of the same data (0.5512 to 0.5883). Architecture thus contributes more than data augmentation alone.

\paragraph{Cross-domain augmentation.} The Jigsaw-transferred samples provide 16,225 additional toxic examples across Labels 1--4 (Table~\ref{tab:data-composition}), broadening the model's exposure to diverse toxic language patterns beyond the gaming domain. The M1 to M1+Jigsaw comparison (+2.6 F1) confirms that this cross-domain transfer provides meaningful gains even with a simple Focal Loss classifier. The improvement is particularly impactful for Labels 3 and 4, which grow from 279 and 60 samples to 1,586 and 538 respectively. A concrete illustration of the domain gap: a Jigsaw threat tends to be syntactically intact (e.g., \textit{``I know where you live and I will make you pay''}), whereas a GameTox threat is fragmented and obfuscated (e.g., \textit{``hope ur family die in fire''}, see Table~\ref{tab:label-taxonomy}); transferred samples therefore broaden lexical coverage but do not fully replicate gaming-register obfuscation patterns.

\paragraph{Rationale-enriched training.} The Qwen2.5-14B rationales supply explicit linguistic reasoning during training, including keyword identification, intent analysis, and domain context. Concatenating rationales with input messages allows the student encoder to associate surface-level toxic patterns with deeper semantic cues during training; rationales are withheld at inference to avoid imposing a teacher dependency at deployment time. This follows the spirit of learning with privileged information, where auxiliary supervision shapes representations that persist at test time even without that context. We acknowledge that this introduces a train/test input distribution shift, and that the contribution of rationale distillation cannot be isolated from SupCon and the rare-class heads in the current ablation (see Limitations).

\paragraph{Rare-class specialisation.} The dedicated binary heads for Labels 4 and 5 (weighted $2\times$) and Supervised Contrastive Loss work in tandem on the shared encoder. The binary heads push the encoder toward features that separate the rarest classes, while the contrastive loss pulls same-class embeddings into tighter clusters.

The M1+Jigsaw to RAKSHAK comparison (+3.7 F1) isolates the combined effect of rationale distillation, contrastive loss, and rare-class heads. The accuracy gap between these two systems (0.8930 vs. 0.9031) further suggests that the multi-task training helps prevent collapse toward the dominant non-toxic class beyond what augmented data alone achieves.

\section{Conclusion}

We presented two systems for the GameTox Shared Task at EEUCA 2026. Our primary system, RAKSHAK, combines multi-task DeBERTa-v3-base training with rationale distillation from Qwen2.5-14B, Supervised Contrastive Loss, dedicated rare-class binary heads, Jigsaw cross-domain transfer, and LLM-generated extremism samples. RAKSHAK achieved a Macro F1 of 0.5883, ranking 7th out of 35 teams. A three-way comparison (M1, M1+Jigsaw, RAKSHAK) shows that Jigsaw transfer contributes +2.6 F1 points while the multi-task architecture adds a further +3.7 points, confirming that the multi-task design contributes more than data augmentation alone.

Three takeaways emerge for toxicity classification under extreme class imbalance: (1)~cross-domain transfer from existing toxicity datasets such as Jigsaw can supplement scarce in-domain data when a reasonable label mapping exists, (2)~concatenating teacher-generated rationales with training inputs, following the distill-then-train paradigm \citep{hsieh2023distilling}, provides a simple mechanism for transferring reasoning from a large model to a smaller encoder without requiring the teacher at inference, and (3)~auxiliary training heads for rare classes combined with Supervised Contrastive Loss can shape the shared representation in ways that benefit the primary classifier.

Future work will focus on finer-grained ablations to isolate the individual contributions of rationale distillation, contrastive loss, and rare-class binary heads within the RAKSHAK architecture. We also plan to explore multilingual encoders such as mDeBERTa-v3 or XLM-R to better capture the non-English utterances present in the GameTox corpus, and to investigate uncertainty-based sample selection for directing rationale generation toward the samples where the model is least confident.

\section*{Limitations}

Our work has several limitations. First, the Jigsaw dataset originates from social media, introducing a domain gap relative to in-game chat; the transferred samples lack the gaming-specific vocabulary, obfuscation, and register typical of World of Tanks communication, and the extent to which social media toxicity patterns transfer to gaming contexts remains an open question. Second, the 100 LLM-generated extremism samples are syntactically cleaner than authentic game chat and were not formally verified for exact-string overlap with the official test set. Third, the M1+Jigsaw to RAKSHAK comparison isolates the combined architectural contribution but does not ablate individual components (Supervised Contrastive Loss, rare-class heads, rationale augmentation) separately; determining which contributes most remains open. The SupCon loss is additionally constrained by the small batch size (32), which limits the frequency of rare-class within-batch pairs; memory-bank approaches or larger batches may yield stronger contrastive signal. Fourth, DeBERTa-v3-base is primarily English-trained, which may limit performance on the substantial non-English content (Polish, Russian, German, and others) present in the corpus. Fifth, concatenating rationales at training time but withholding them at inference introduces an input distribution shift: the encoder is optimised on \texttt{[MESSAGE] [SEP] [RATIONALE]} but evaluated on \texttt{[MESSAGE]} alone. Although this follows the learning-with-privileged-information paradigm, it means the contribution of rationale distillation to the overall gain cannot be cleanly attributed, and the Supervised Contrastive Loss and rare-class heads may account for a larger share. More principled alternatives, such as an auxiliary rationale prediction head or KL-divergence matching against the teacher's output distribution, would avoid this shift in future work. Sixth, we report only macro-level aggregate metrics; per-class F1 for rare labels (particularly Labels 4 and 5) would provide a more transparent view of where the system's gains are concentrated.

\section*{Acknowledgments}

We thank the organisers of the EEUCA 2026 shared task for providing the dataset and evaluation infrastructure.

\bibliography{custom}

\appendix

\section{Prompt Template for Extremism Sample Generation}
\label{sec:prompt}

The following prompt was used with Qwen2.5-14B to generate synthetic extremism training samples (Section~\ref{sec:extremism-aug}). Extremist keywords are replaced with placeholder tokens to work around safety refusal.

\begin{quote}
\small
\texttt{You are helping create training data for an academic shared task on toxicity classification in online gaming. The task is called GameTox, organised as part of the EEUCA workshop at ACL 2026. It classifies chat messages from the online multiplayer game World of Tanks into six toxicity categories.}

\medskip

\texttt{Label 5 (Extremism) is defined as: messages containing extremist content, radicalisation, incitement to ideological violence, glorification of hate groups, or promotion of radical ideologies, as they appear in online game chat. This includes references to real-world extremist movements, political radicalisation, and calls for violence framed within the gaming context.}

\medskip

\texttt{Generate 5 short in-game chat messages (1--2 sentences each) that would be classified as Label 5. Messages should read like real game chat: informal, possibly containing typos, abbreviations, or mixed languages. Use the placeholder tokens [WORD1], [WORD2], and [WORD3] where extremist or radical terms would naturally appear. Do not use any actual slurs or extremist language yourself.}

\medskip

\texttt{Output only the messages, one per line, with no numbering or extra commentary.}
\end{quote}

After generation, placeholder tokens are replaced with real extremist keywords obtained through the keyword mining and expansion steps described in Section~\ref{sec:extremism-aug}.

\end{document}